\documentclass[10pt,twocolumn,letterpaper]{article}

\usepackage[pagenumbers]{cvpr}

\definecolor{cvprblue}{rgb}{0.21,0.49,0.74}
\usepackage[breaklinks,colorlinks,allcolors=cvprblue]{hyperref}
\hypersetup{
  pdftitle={EgoMIP: Learning Multi-Level Interaction Priors for Robot Manipulation},
  pdfsubject={EgoMIP: egocentric world priors for robot manipulation},
  pdfauthor={Bowen Yang, Ruizhe Yang, Xinke Lyu, Zhongzhong Liu, Xin Qiu, Xu-Yao Zhang, Zhicheng Wang, Jian Xu},
  pdfkeywords={robot manipulation, egocentric video, world-action models, contact learning}
}

\title{EgoMIP: Learning Multi-Level Interaction Priors for Robot Manipulation}
\author{
    {\large
    \textbf{Bowen Yang}$^{1*\ddagger}$ \quad
    \textbf{Ruizhe Yang}$^{2*}$ \quad
    \textbf{Xinke Lyu}$^{3}$ \quad
    \textbf{Zhongzhong Liu}$^{4}$} \\[-0.1em]
    {\large
    \textbf{Xin Qiu}$^{5}$ \quad
    \textbf{Xu-Yao Zhang}$^{1}$ \quad
    \textbf{Zhicheng Wang}$^{5\dagger}$ \quad
    \textbf{Jian Xu}$^{1\dagger}$} \\[0.35em]
    {\large
    $^{1}$Institute of Automation, Chinese Academy of Sciences}
    \\[-0.1em]
    {\large
    $^{2}$Harbin Institute of Technology \quad
    $^{3}$Beijing University of Posts and Telecommunications}
    \\[-0.1em]
    {\large
    $^{4}$University of the Chinese Academy of Sciences \quad
    $^{5}$Powertech Inc.}
}

\begin{document}
\raggedbottom
\maketitle

\begingroup
\renewcommand{\thefootnote}{\fnsymbol{footnote}}

\footnotetext[1]{Equal contribution.}

\footnotetext[2]{Corresponding authors: Zhicheng Wang and Jian Xu.
}

\footnotetext[3]{This work was conducted during an internship
at Powertech Inc.}

\endgroup

\begin{abstract}
Human egocentric videos provide rich observations of object interactions and environmental changes, yet embodiment differences make such experience difficult to transfer directly into robotic control. We introduce EgoMIP, a unified framework that learns multi-level interaction priors by combining transferable visual foresight with embodiment-specific contact learning. Leveraging public and in-house egocentric videos, EgoMIP constructs interaction-aware world supervision through task alignment, camera-motion compensation, interaction-region masking, and future visual representation learning. Our compact World--Contact--Action (WCA) transformer integrates three specialized experts: World predicts future visual dynamics, Contact models robot-specific 2.5D interaction trajectories, and Action generates executable robot actions. Human videos supervise world representations, while contact and action learning rely exclusively on robot demonstrations. On 50 RoboTwin 2.0 manipulation tasks, EgoMIP achieves 93.36\% success rate in clean scenes and 93.34\% in randomized scenes with egocentric world supervision, improving generalization over the robot-only baseline. The benefits are particularly pronounced in task-aligned settings, where egocentric supervision yields an 8.2 percentage-point improvement in randomized success on a targeted 10-task evaluation. Real-world experiments on a  dual-arm robot further indicate promising performance in contact-intensive real-world manipulation. These results highlight the value of integrating shared world foresight with embodiment-specific contact priors for robust robotic manipulation.
\end{abstract}

\section{Introduction}
\label{sec:intro}

Robotic manipulation requires more than recognizing objects or producing a sequence of motor commands. A robot must anticipate how an interaction will change the environment~\cite{visual_foresight,unipi,vlp}, determine where and how to establish contact~\cite{hierarchical_diffusion_policy}, and translate this understanding into executable actions~\cite{diffusion_policy,act}. These requirements operate at different levels of abstraction: future scene changes describe what should happen, contact motion specifies how the robot should interact, and joint-space control determines how the interaction is executed. Connecting these complementary levels is essential for reliable manipulation, particularly in tasks involving precise physical interactions and multiple execution stages.

\begin{figure*}[t]
    \centering
    \includegraphics[
        width=0.85\textwidth,
        height=0.50\textheight,
        keepaspectratio
    ]{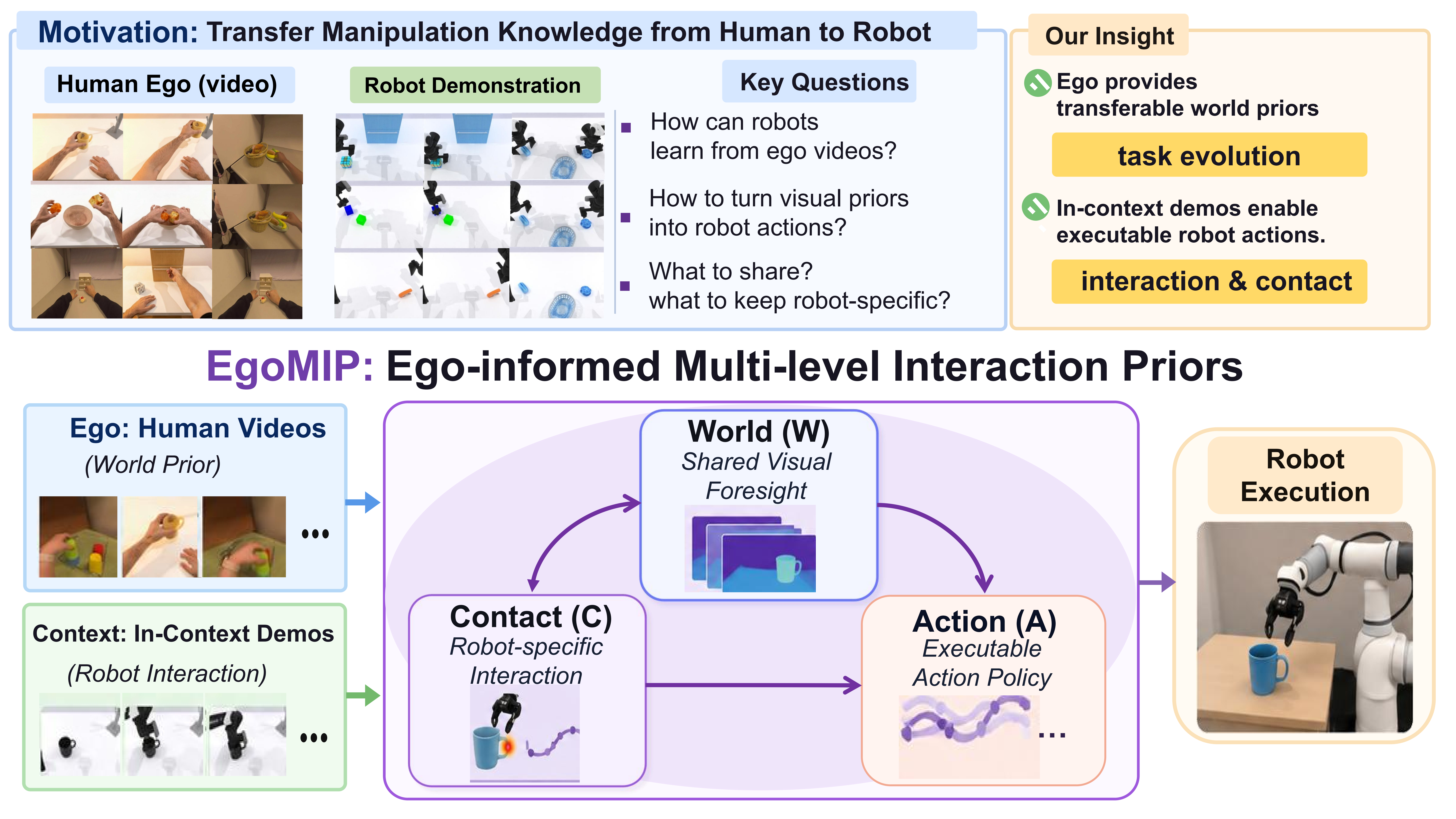}
    \vspace{-1mm}
    \caption{
        \textbf{Motivation and overview of EgoMIP.}
        Human egocentric videos provide transferable world priors,
        while robot demonstrations provide embodiment-specific
        contact information for executable manipulation.
    }
    \label{fig:motivation}
    \vspace{-2mm}
\end{figure*}

Recent vision-language-action (VLA)~\cite{openvla,pi0} and world--action models~\cite{gr1,vpp} have advanced robotic manipulation from visual observations and demonstrations~\cite{rt1,octo}. Nevertheless, collecting diverse robot trajectories remains costly~\cite{droid,bridgedata_v2,open_x_embodiment}, and even strong policies may struggle when object configurations, backgrounds, or interaction dynamics differ from those encountered during training~\cite{pi05,robotwin2}. Human egocentric videos offer a complementary source of experience: they record how objects are moved, transformed, and used across varied environments without requiring robot teleoperation~\cite{ego4d,epic_kitchens,egodex}. Yet their value for robot learning should not be confused with direct action transfer. Human hands and robot grippers can bring about similar changes in the world while relying on different contact geometries and control spaces~\cite{whirl,mimicplay}.

Despite this opportunity, learning manipulation policies from egocentric experience presents three challenges. First, camera motion, hand occlusion, and background variation complicate the extraction of transferable world knowledge~\cite{human_video_transfer,egowam}, potentially causing future prediction to capture viewpoint changes rather than task-relevant object dynamics. Second, visual foresight alone does not specify robot-specific contact trajectories, approach directions, or gripper states. Directly using human hand trajectories as robot contact targets can further introduce embodiment-dependent errors~\cite{egoforesight}. Third, contact errors may accumulate during closed-loop, multi-stage execution, particularly under scene randomization and real-world disturbances. Therefore, human-derived visual knowledge must be grounded in robot-specific interaction geometry to reliably support action generation.

Existing studies have explored learning from egocentric demonstrations through affordance prediction~\cite{li2024preciseaffordances,human_video_transfer}, human motion alignment~\cite{skelwam}, and future-outcome supervision~\cite{egoforesight}. In contrast, we investigate a deliberately asymmetric interface between the two embodiments. Rather than requiring human videos to provide robot-compatible contact or action labels, we use them to supervise shared visual foresight, while learning embodiment-specific contact motion and executable actions from robot demonstrations. Our central premise is that the consequences of interaction can be more transferable than the physical movements that produce them.

To this end, we introduce EgoMIP, a unified World--Contact--Action (WCA) framework for learning multi-level interaction priors. EgoMIP constructs world-level supervision from task-aligned public and in-house egocentric videos through camera-motion compensation, hand and arm masking, and future visual feature extraction using a frozen visual encoder~\cite{dinov3}. Human videos thus supervise task-relevant visual changes rather than robot joint control, while robot observations contribute to world representation learning in the same future-feature space.

On the robot side, EgoMIP introduces a robot-specific 2.5D contact representation between visual foresight and action generation. A compact WCA transformer jointly models three specialized experts: World predicts future visual representations, Contact models bimanual interaction trajectories, and Action generates executable 14-dimensional robot action chunks. Human videos supervise only World, while robot demonstrations supervise all three experts. Rather than forming a rigid sequential pipeline, the experts jointly learn complementary representations. A second training stage further refines the policy through temporal-prior adaptation and auxiliary far-motion prediction.

We evaluate EgoMIP on 50 RoboTwin~2.0 manipulation tasks under clean and randomized settings~\cite{robotwin2}. The two-stage WCA policy achieves 93.36\% success in clean scenes and 93.34\% in randomized scenes. In a separate limited-demonstration study on ten task-aligned tasks, matched human World-only supervision improves randomized-scene success by 8.2 percentage points, although its benefits depend on the evaluation setting and supervision interface. We further examine physical feasibility through preliminary deployment on a dual-arm robot.
In summary, our contributions are threefold:
\vspace{-3pt}
\begingroup
\setlength{\topsep}{0pt}
\setlength{\itemsep}{0pt}
\setlength{\parsep}{0pt}
\setlength{\partopsep}{0pt}
\begin{itemize}
    \item \textbf{Interaction-Oriented Egocentric World Supervision.}
    We develop a task-aligned pipeline to extract camera-compensated,
    agent-masked future visual representations from human videos,
    enabling world-level supervision without robot action or contact labels.

    \item \textbf{Asymmetric World--Contact--Action Learning.}
    We introduce EgoMIP, a lightweight framework that jointly learns
    visual foresight, robot-specific 2.5D contact motion, and executable
    actions through asymmetric human--robot supervision.

    \item \textbf{Multi-Task Evaluation and Real-World Deployment.}
    We evaluate EgoMIP on 50 RoboTwin~2.0 tasks, analyze component
    contributions and task-level supervision effects, and demonstrate
    preliminary deployment on a physical dual-arm robot.
\end{itemize}
\endgroup
\vspace{-4pt}

\section{Related Work}
\label{sec:related_work}

\noindent\textbf{Learning Transferable Priors from Egocentric Videos.}
Human egocentric videos provide scalable interaction knowledge for robot learning, capturing diverse object manipulations and scene transitions~\cite{ego4d,epic_kitchens,egodex}. Existing studies transfer such experience through human motion reconstruction and alignment~\cite{skelwam}, action retargeting and reliability-aware supervision~\cite{egoforesight}, latent action learning~\cite{lapa}, and future visual prediction or cross-demonstration correspondence~\cite{watch,egowam}. Despite these advances, directly transferring human motion or contact information remains challenging due to embodiment differences and the heterogeneous visual conditions of egocentric recordings.

In contrast, EgoMIP focuses on transferable \emph{world-level interaction priors} rather than human-motion retargeting. Human videos supervise shared visual foresight, while robot demonstrations provide embodiment-specific contact geometry and executable action supervision. This separation enables cross-embodiment knowledge transfer without imposing human-specific motion constraints on robot control.

\noindent\textbf{World Modeling and Contact-Grounded Robotic Manipulation.}
Vision-language-action models map visual observations and task instructions to robot actions~\cite{openvla,pi0,robot_vla}, while world--action models additionally predict future environment states to guide manipulation~\cite{gr1,vpp,lilawam,gawm}. Representative approaches include latent world--action reasoning~\cite{lilawam}, multi-horizon future visual prediction~\cite{gawm}, and predictive representation learning~\cite{vjepa2}. However, visual foresight alone does not specify the contact geometry, approach direction, or gripper motion required for precise physical interaction.

Robot interaction modeling has explored contact-guided trajectory generation~\cite{hierarchical_diffusion_policy}, geometric action representations~\cite{geometric_action_model,peract,dp3}, and structured motion prediction~\cite{skelwam}. Nevertheless, integrating transferable human visual knowledge with robot-specific contact supervision remains challenging. EgoMIP addresses this gap through a unified \emph{World--Contact--Action} framework, jointly learning shared future visual dynamics, robot-specific 2.5D contact trajectories, and executable actions without requiring human-to-robot contact correspondence.

\section{Method}
\label{sec:method}

\begin{figure*}[t]
    \centering
    \includegraphics[width=0.98\textwidth]{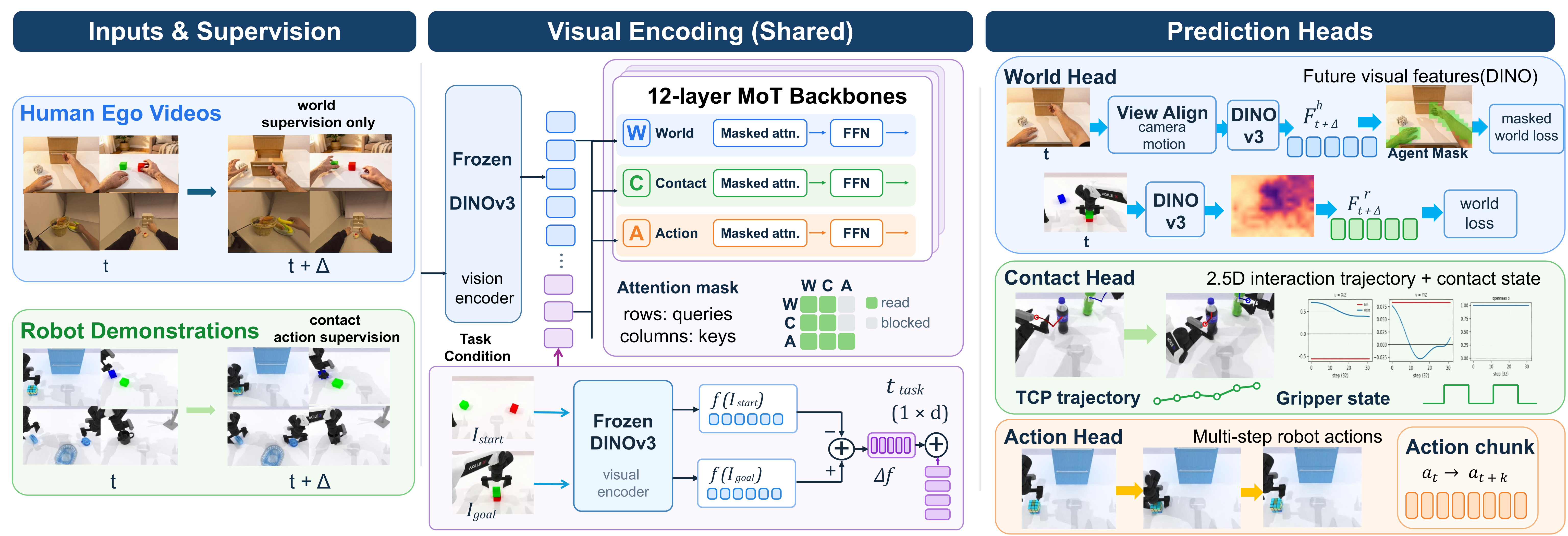}
    \caption{
        \textbf{Architecture of EgoMIP.}
        Human egocentric videos and robot demonstrations
        jointly provide future visual feature supervision,
        with camera alignment and agent masking applied
        to human world supervision.
        Robot demonstrations additionally supervise
        2.5D contact trajectories and executable actions.
        A unified World--Contact--Action Transformer
        integrates these complementary signals through
        directed attention and demonstration-derived
        visual task conditioning.
    }
    \label{fig:egomip_architecture}
\end{figure*}

\subsection{Egocentric World Foresight Learning}
\label{sec:ego_world}

Human egocentric videos provide transferable knowledge of object interactions and environmental changes~\cite{r3m,vip,vc1,mvp}. However, camera motion, hand occlusion, and appearance variations complicate direct learning from raw videos. We therefore construct an interaction-aware pipeline to extract task-relevant future visual representations.

\noindent\textbf{Task-Aligned Egocentric Video Processing.}
We combine public human interaction videos from EgoDex and TACO~\cite{egodex,taco}, synthetic manipulation videos from HumanGen~\cite{zhou2026zerowam}, and in-house human-object interaction videos. Video segments are aligned with robot task categories to construct task-conditioned samples. Given $\mathcal{V}^{h}=\{I_t^{h}\}_{t=1}^{T}$, we sample observation--future pairs $(I_t^{h},I_{t+\Delta}^{h})$. To reduce viewpoint variations, future observations are warped into the reference view through camera-motion compensation. Hand and forearm regions, together with invalid warped regions, are masked to suppress embodiment-specific visual distractions.

\noindent\textbf{Future Visual Representation Supervision.}
Rather than reconstructing future RGB frames, we use a frozen DINOv3 encoder~\cite{dinov3} to extract future visual targets. Given the aligned observation $\widetilde{I}_{t+\Delta}^{h}$, the target is

\begin{equation}
    \mathbf{z}_{t+\Delta}^{h}
    = \phi(\widetilde{I}_{t+\Delta}^{h}).
    \label{eq:ego_world_target}
\end{equation}

Given the current observation and task condition, the World expert predicts the corresponding future representation $\hat{\mathbf{z}}_{t+\Delta}^{h}$. We optimize a masked feature prediction objective:
\begin{equation}
    \mathcal{L}_{W}^{h}
    =
    \frac{1}{\sum_{p}m_p}
    \sum_{p}m_p w_p
    \left(
    1-
    \frac{
    \langle\hat{\mathbf{z}}_p^{h},
    \mathbf{z}_p^{h}\rangle
    }{
    \|\hat{\mathbf{z}}_p^{h}\|
    \|\mathbf{z}_p^{h}\|
    }
    \right),
    \label{eq:ego_world_loss}
\end{equation}
where $p$ indexes visual tokens, $m_p$ excludes masked agent regions and invalid warped regions, and $w_p$ denotes an optional interaction-effect weight ($w_p=1$ for uniform supervision). Robot observations share the same future-feature space, enabling complementary world-level supervision. Human videos provide no contact or robot-action targets: these objectives are supervised exclusively by robot demonstrations.

\subsection{Contact-Grounded Interaction Modeling}
\label{sec:contact_modeling}

While visual foresight captures how the environment is expected to evolve, it does not explicitly specify the physical interactions required to achieve the predicted changes. In particular, precise manipulation depends on where the robot establishes contact, how its end-effector moves, and when the gripper opens or closes. To bridge the gap between visual prediction and executable control, we introduce a structured 2.5D contact representation that captures robot-specific interaction geometry and temporal motion.

\noindent\textbf{Robot-Centric Contact Representation.}
Given a robot demonstration, we approximate the interaction point using the end-effector pose and a calibrated tool-center-point (TCP) offset:
\begin{equation}
    \mathbf{p}_{t}^{a}
    =
    \mathbf{t}_{t}^{a}
    +
    \mathbf{R}_{t}^{a}\mathbf{d},
    \label{eq:contact_point}
\end{equation}
where $\mathbf{t}_{t}^{a}$ and $\mathbf{R}_{t}^{a}$ denote the end-effector position and orientation of arm $a$, and $\mathbf{d}$ is the fixed offset to the interaction point. We transform the resulting points into the reference camera coordinate system and represent each future contact state as
\begin{equation}
    \mathbf{c}_{t+i}^{a}
    =
    \left[
    \frac{X_{t+i}^{a}}{Z_{t+i}^{a}},
    \frac{Y_{t+i}^{a}}{Z_{t+i}^{a}},
    \log\frac{Z_{t+i}^{a}}{Z_t^{a}},
    o_{t+i}^{a}
    \right],
    \label{eq:contact_representation}
\end{equation}
where $(X,Y,Z)$ denotes the camera-frame interaction point and $o\in[0,1]$ represents the normalized gripper openness. This representation combines image-plane contact location, relative depth, and gripper state while avoiding dependence on absolute scene scale.

\noindent\textbf{Structured Contact Motion Modeling.}
We model the bimanual contact motion on a future horizon of $H=32$ steps:
\begin{equation}
    \mathbf{C}_{t}
    =
    \left[
    \mathbf{c}_{t+1},\ldots,\mathbf{c}_{t+H}
    \right]
    \in \mathbb{R}^{H\times 8},
    \label{eq:contact_trajectory}
\end{equation}
where each step concatenates the four-dimensional contact states of both arms. A dimension-wise validity mask handles invalid observations, and the resulting trajectories supervise the Contact expert through flow matching~\cite{flowmatching}.

Importantly, contact supervision is derived exclusively from robot demonstrations, without requiring human hand-to-robot contact correspondence. In this way, EgoMIP separates transferable world-level knowledge from embodiment-specific interaction geometry, allowing contact representations to provide an intermediate grounding between visual foresight and robot action generation.

\subsection{Unified World--Contact--Action Learning}
\label{sec:wca_learning}

Building upon the transferable world supervision in Sec.~\ref{sec:ego_world} and the robot-specific contact representation in Sec.~\ref{sec:contact_modeling}, we introduce a unified World--Contact--Action (WCA) architecture that integrates interaction priors at different levels of abstraction. Rather than treating visual prediction and action generation as independent objectives, EgoMIP jointly models future world dynamics, structured contact motion, and executable robot actions within a compact transformer.

\noindent\textbf{Multi-Level Expert Architecture.}
Given the current visual observation, the proprioceptive state of the robot, and the task condition, a frozen visual encoder extracts compact visual representations through 64 learnable visual queries. These features are combined with modality-specific and task-conditioning tokens and processed by a 12-layer transformer~\cite{transformer} with a hidden dimension of 768 and 8 attention heads.

The architecture consists of three specialized experts. The \emph{World} expert predicts future visual representations, capturing anticipated changes in the environment. The \emph{Contact} expert models the 32-step bimanual contact trajectory using 16 motion tokens and a contact-state token. The \emph{Action} expert takes robot proprioception and 32 action tokens to generate a sequence of 14-dimensional robot actions. Experts maintain modality-specific representations while exchanging information through masked attention.

\noindent\textbf{Integration of Cross-Level Information.}
To preserve the complementary roles of different interaction priors, we introduce an asymmetric attention structure. The World and Contact experts do not attend to Action tokens, whereas the Action expert can attend to World, Contact, and Action representations. This allows action prediction to incorporate both anticipated visual changes and structured interaction geometry without making world or contact prediction dependent on the action-generation tokens.

Moreover, we randomly remove contact conditioning from the Action expert during training to reduce its dependence on explicit contact features. We also include an auxiliary training mode that optimizes World and Contact prediction without action supervision. These mechanisms encourage the model to learn complementary interaction representations while maintaining robust action generation.

\noindent\textbf{Asymmetric Supervision and Joint Optimization.}
EgoMIP adopts an asymmetric supervision strategy for human and robot demonstrations. For robot samples, all three experts are jointly supervised, including future visual prediction, contact trajectory prediction, and action generation. In particular, the robot World expert is trained with uniform feature supervision over valid visual regions, while the Contact and Action experts are optimized using flow-matching objectives~\cite{flowmatching}:

\begin{equation}
    \mathcal{L}_{R}
    =
    \mathcal{L}_{W}^{R}
    +
    \mathcal{L}_{C}^{R}
    +
    \mathcal{L}_{A}^{R},
    \label{eq:robot_wca_loss}
\end{equation}

where $\mathcal{L}_{W}^{R}$ denotes the uniform future visual feature prediction loss, and $\mathcal{L}_{C}^{R}$ and $\mathcal{L}_{A}^{R}$ represent the flow-matching objectives for robot contact trajectories and actions, respectively.

For human egocentric videos, only the World expert receives supervision. Unlike the uniform visual supervision used for robot observations, human world supervision employs an adaptive action-effect feature (AAEF) objective to emphasize visually informative regions associated with interaction-induced scene changes.

\begin{equation}
    \mathcal{L}_{H}
    =
    \mathcal{L}_{\mathrm{AAEF}}^{H}.
    \label{eq:human_wca_loss}
\end{equation}

\noindent\textbf{Adaptive Action-Effect Feature Supervision.}
After camera-motion compensation and agent-region
masking, we compute the feature-change magnitude of each
valid token as a proxy for interaction-related visual
changes:
\begin{equation}
    \delta_p
    =
    1-
    \cos\left(
    \mathbf{z}_{t+\Delta,p}^{H},
    \mathbf{z}_{t,p}^{H}
    \right),
    \label{eq:effect_magnitude}
\end{equation}

Here, the feature-change magnitude serves as a heuristic
proxy for visual changes rather than a causal estimate
of action effects. We define the adaptive weight as

\begin{equation}
    w_p
    =
    v_p(1-a_p)
    \left(
    1+\beta\frac{\delta_p}{\bar{\delta}+\epsilon}
    \right),
    \label{eq:adaptive_weight}
\end{equation}

where $v_p$ indicates whether the visual token is valid after geometric alignment, $a_p$ denotes the agent-region mask, $\bar{\delta}$ is the mean interaction-effect magnitude over valid regions, $\beta$ controls the effect-aware weighting strength, and $\epsilon$ is a small constant for numerical stability.

The resulting AAEF objective is formulated as

\begin{equation}
    \mathcal{L}_{\mathrm{AAEF}}^{H}
    =
    \frac{
    \sum_p w_p
    \left(
    1-
    \cos(
    \hat{\mathbf{z}}_{t+\Delta,p}^{H},
    \mathbf{z}_{t+\Delta,p}^{H}
    )
    \right)
    }{
    \sum_p w_p
    }.
    \label{eq:human_aaef_loss}
\end{equation}

AAEF reweights only valid, unmasked tokens in human
World supervision. Robot World prediction retains
uniform weighting, and human videos provide no Contact
or Action targets.

The overall training objective is defined as

\begin{equation}
    \mathcal{L}_{\mathrm{WCA}}
    =
    \mathcal{L}_{R}
    +
    \alpha\mathcal{L}_{H},
    \label{eq:joint_wca_loss}
\end{equation}

where $\alpha$ controls the relative contribution of human egocentric supervision. This asymmetric design enables human and robot demonstrations to share predictive visual knowledge while keeping contact and action learning grounded in robot-specific supervision.

\noindent\textbf{Two-Stage Policy Learning.}
We adopt a two-stage training strategy for high-performance manipulation. The first stage jointly learns world prediction, contact motion, and action generation through the WCA objectives. In the second stage, we introduce a lightweight temporal prior adapter and an auxiliary far-motion prediction head to further refine the learned policy. The additional temporal supervision complements the original contact and action objectives while preserving the unified WCA structure. During inference, the policy predicts executable robot action chunks from current observations and task conditions, without requiring human egocentric demonstrations as online input.

\section{Experiments}
\label{sec:experiments}

In this section, we comprehensively evaluate EgoMIP from three perspectives: multi-task manipulation performance, the effectiveness of egocentric world supervision, and the contribution of multi-level interaction priors. We first compare EgoMIP with representative VLA and world--action models on 50 RoboTwin 2.0 manipulation tasks under both clean and randomized settings. We then conduct ablation studies to investigate the individual contributions of egocentric visual foresight, robot-specific contact modeling, and unified World--Contact--Action learning. In addition, we analyze the generalization benefits of egocentric supervision in different data and task categories. Finally, we investigate real-world deployment on a Galaxea dual-arm robot, focusing on contact-intensive manipulation and multi-stage execution. Additional real-world evaluation details, ablation studies, and contact trajectory analysis are provided in Appendices~\ref{sec:appendix_realworld}--\ref{sec:appendix_extended}.

\subsection{Multi-Task Robot Manipulation}
\label{sec:multi_task}

\noindent\textbf{Evaluation Protocol.}
We evaluate EgoMIP on RoboTwin 2.0~\cite{robotwin2}, a benchmark consisting of 50 different bimanual manipulation tasks, including object grasping, placement, stacking, and contact-intensive interactions. The robot training set contains 27,071 demonstrations, covering both clean and randomized environments. Egocentric videos provide additional world-level supervision, while contact and action prediction are supervised by robot demonstrations. We evaluate a unified multi-task policy under clean and randomized settings using fixed evaluation seeds, with up to 50 trials per task. All simulation results for our model variants are
averaged over three training seeds.

\noindent\textbf{Comparison with Existing Methods.}
Table~\ref{tab:robotwin_main} presents a comparison with representative Vision-Language-Action (VLA) and World-Action Models (WAMs). EgoMIP achieves \textbf{93.36\%} success in clean scenes and \textbf{93.34\%} in randomized scenes, demonstrating competitive multi-task manipulation performance. These results numerically exceed the reported performance of several established world-action models, including Motus, LaWAM, Fast-WAM, and LingBot-VA. In particular, EgoMIP achieves approximately 1.5 percentage points higher success than Fast-WAM in both evaluation settings.

Unlike large video-generation-based world-action models, EgoMIP employs a compact WCA backbone with approximately 0.31B parameters, excluding the frozen visual encoder. By learning future visual representations rather than generating dense RGB videos, our framework offers a lightweight approach to integrating world foresight and physical interaction learning. This demonstrates a favorable performance-capacity trade-off relative to several multi-billion-parameter world-action architectures.

\begin{table}[t]
\centering
\caption{Performance comparison on RoboTwin 2.0-Full. Bold scores indicate the
best performance.}
\vspace{-2mm}
\label{tab:robotwin_main}
\small
\renewcommand{\arraystretch}{1.12}
\begin{tabular*}{\columnwidth}{@{\extracolsep{\fill}}lcccc@{}}
\hline
\textbf{Method} & \textbf{Type}
& \textbf{Clean} & \textbf{Rand.} & \textbf{Avg.} \\
\hline
X-VLA~\cite{xvla}
    & VLA & 72.80 & 72.84 & 72.82 \\
$\pi_{0.5}$~\cite{pi05}
    & VLA & 82.74 & 76.76 & 79.75 \\
\hline
Motus~\cite{motus}
    & WAM & 88.66 & 87.02 & 87.84 \\
LaWAM~\cite{lawam}
    & WAM & 92.64 & 89.80 & 91.22 \\
Fast-WAM~\cite{fastwam}
    & WAM & 91.88 & 91.78 & 91.83 \\
LingBot-VA~\cite{lingbotva}
    & WAM & 92.90 & 91.50 & 92.20 \\
\hline
\textbf{EgoMIP (Ours)}
    & WAM & \textbf{93.36} & \textbf{93.34}
    & \textbf{93.35} \\
\hline
\end{tabular*}
\end{table}

\subsection{Egocentric Supervision and Generalization}
\label{sec:ego_generalization}

We investigate how egocentric visual supervision and robot-specific contact modeling contribute to manipulation performance. Table~\ref{tab:egomip_ablation} compares five configurations on all 50 RoboTwin 2.0 tasks, varying the inclusion of the Contact expert and the allocation of human supervision. All configurations retain robot-supervised world prediction and action learning. We also examined a separate task-aligned study with limited robot demonstrations to characterize the effects of human experience under a limited training budget.

\noindent\textbf{Effect of Robot Contact Modeling.}
We first compare world action learning with and without explicit contact modeling. Under robot-only supervision, the introduction of the Contact expert increases the clean-scene success from 90.28\% to 93.06\% and the randomized-scene success from 89.16\% to 92.04\%, corresponding to improvements of 2.78 and 2.88 percentage points, respectively. Contact modeling also provides gains when egocentric world supervision is present: the full model exceeds its no-Contact counterpart by 2.82 points in clean scenes and 3.49 points in randomized scenes. These comparisons support the use of structured 2.5D TCP trajectories and gripper states as intermediate interaction-motion cues for action generation.

\noindent\textbf{Effect of Egocentric World Supervision.}
We first examine the contribution of human world supervision under different model configurations. Without the Contact expert, adding Ego World supervision improves clean and randomized success by 0.26 and 0.69 percentage points, respectively. With robot-specific contact modeling, the full EgoMIP further improves the success from 93.06\% to 93.36\% in clean scenes and from 92.04\% to 93.34\% in randomized scenes, corresponding to gains of 0.30 and 1.30 percentage points. In particular, the larger gain under randomization suggests that transferable visual foresight is particularly beneficial when the scene configuration varies. These consistent improvements across both configurations support the complementary role of human-derived world knowledge in robot manipulation.

\noindent\textbf{Asymmetric Supervision Allocation.}

We further investigate whether human videos should supervise contact motion in addition to world representations. Adding human contact supervision to the robot-only WCA model yields 92.92\% clean and 92.19\% randomized success, compared with 93.06\% and 92.04\% under robot-only supervision. This results in a marginal decrease of 0.14 percentage points in clean scenes and a negligible gain of 0.15 points under randomization. In contrast, EgoMIP, which restricts human supervision to the World while retaining robot-specific Contact and Action supervision, achieves 93.36\% and 93.34\%, respectively. These results suggest that human contact supervision provides no consistent additional benefit, whereas selectively transferring visual foresight while preserving robot-specific interaction supervision yields the strongest overall performance.

\noindent\textbf{Effect of Human Supervision Allocation under Limited Robot Demonstrations.}
We conducted a controlled study on ten task-aligned RoboTwin tasks, using approximately 150 robot demonstrations per task, to compare different human supervision strategies under a limited robot-data budget. As shown in Table~\ref{tab:ego_generalization}, adding matched human World-only supervision improves success from 72.0\% to 76.2\% in clean scenes and from 60.2\% to 68.4\% in randomized scenes, producing gains of 4.2 and 8.2 percentage points, respectively, and a 6.2-point improvement in average success. In contrast, joint human World--Contact supervision achieves 71.6\% clean and 61.4\% randomized success, yielding only a marginal average gain of 0.4 percentage points over robot-only training. These results indicate that, under limited robot demonstrations, transferring human visual foresight alone is more beneficial than additionally imposing human contact supervision. This controlled comparison further supports EgoMIP's asymmetric supervision strategy: sharing world-level knowledge across embodiments while retaining robot-specific contact and action supervision.

\begin{table*}[t]
\centering
\caption{
Component and supervision ablations on the
50 RoboTwin 2.0 tasks.
All values are success rates (\%).
Avg. denotes the arithmetic mean of Clean and Randomized success rates.
}
\vspace{-2mm}
\label{tab:egomip_ablation}
\small
\setlength{\tabcolsep}{5pt}
\renewcommand{\arraystretch}{1.12}
\begin{tabular*}{\textwidth}{
    @{\extracolsep{\fill}}lcccccc@{}
}
\toprule
\textbf{Model Configuration}
& \multicolumn{3}{c}{\textbf{Supervision}}
& \multicolumn{3}{c}{\textbf{Success Rate (\%)}} \\
\cmidrule(lr){2-4}
\cmidrule(lr){5-7}
& \textbf{Robot Contact}
& \textbf{Human World}
& \textbf{Human Contact}
& \textbf{Clean}
& \textbf{Rand.}
& \textbf{Avg.} \\
\midrule

W--A (Robot-only)
& --
& --
& --
& 90.28
& 89.16
& 89.72 \\

W--A + Ego World
& --
& $\checkmark$
& --
& 90.54
& 89.85
& 90.20 \\

W--C--A (Robot-only)
& $\checkmark$
& --
& --
& 93.06
& 92.04
& 92.55 \\

\midrule

EgoMIP + Human Contact
& $\checkmark$
& $\checkmark$
& $\checkmark$
& 92.92
& 92.19
& 92.56 \\

\textbf{EgoMIP (Full)}
& $\checkmark$
& $\checkmark$
& --
& \textbf{93.36}
& \textbf{93.34}
& \textbf{93.35} \\

\bottomrule
\end{tabular*}
\end{table*}

\begin{table}[t]
    \centering
   \caption{
     Controlled comparison of human supervision strategies
     under limited robot demonstrations on ten task-aligned
     RoboTwin 2.0 tasks. W-only and W+C denote human
     World-only and joint World--Contact supervision,
     respectively. All results are success rates (\%).
     }
    \vspace{-2mm}
    \label{tab:ego_generalization}
    \small
    \begin{tabular*}{\columnwidth}{
        @{\extracolsep{\fill}}lccc@{}
    }
        \toprule
        Model Configuration
        & Clean
        & Rand.
        & Avg. \\
        \midrule
        Robot-only W--C--A
        & 72.0
        & 60.2
        & 66.1 \\

        + Matched Ego (W+C)
        & 71.6
        & 61.4
        & 66.5 \\

        + Matched Ego (W-only)
        & \textbf{76.2}
        & \textbf{68.4}
        & \textbf{72.3} \\

        \midrule
        $\Delta$ (W-only vs. Robot-only)
        & $+4.2$
        & $\boldsymbol{+8.2}$
        & $+6.2$ \\
        \bottomrule
    \end{tabular*}
\end{table}

\begin{table}[t]
    \centering
    \caption{
        Task-level success-rate changes with human World-only
        supervision relative to robot-only training. Contact and Action
        remain supervised exclusively by robot demonstrations.
        Values are in percentage points (pp); Avg. denotes the
        mean of Clean and Randomized changes.
    }
    \vspace{-2mm}
    \label{tab:task_level_analysis}
    \footnotesize
    \setlength{\tabcolsep}{3pt}
    \renewcommand{\arraystretch}{1.08}
    \begin{tabular*}{\columnwidth}{
        @{\extracolsep{\fill}}lrrr@{}
    }
        \toprule
        \textbf{Task}
        & \textbf{Clean}
        & \textbf{Rand.}
        & \textbf{Avg.} \\
        \midrule

        \multicolumn{4}{l}{\textit{Object-centric Manipulation}} \\
        \addlinespace[1pt]
        \texttt{stack\_blocks\_three}
        & +34.0 & +36.0 & +35.0 \\
        \texttt{place\_cans\_plasticbox}
        & +24.0 & +28.0 & +26.0 \\
        \texttt{place\_bread\_skillet}
        & +14.0 & +22.0 & +18.0 \\
        \texttt{pick\_dual\_bottles}
        & +20.0 & +12.0 & +16.0 \\

        \midrule
        \multicolumn{4}{l}{\textit{Task-specific Interactions}} \\
        \addlinespace[1pt]
        \texttt{beat\_block\_hammer}
        & -4.0 & -16.0 & -10.0 \\
        \texttt{turn\_switch}
        & 0.0 & -20.0 & -10.0 \\
        \texttt{rotate\_qrcode}
        & -12.0 & -18.0 & -15.0 \\
        \texttt{hanging\_mug}
        & -10.0 & -12.0 & -11.0 \\
        \bottomrule
    \end{tabular*}
    \vspace{-1mm}
\end{table}

\noindent\textbf{Task-level Effects of Human World Supervision.}
We further examine task-level variations when adding human World-only supervision to robot-only training. Contact and Action remain supervised exclusively by robot demonstrations. As shown in Table~\ref{tab:task_level_analysis}, substantial gains are observed in object-centric tasks involving spatial arrangement, object placement, and bimanual grasping, while tasks involving tool use, articulated interactions, and constrained manipulation exhibit less favorable results. These variations indicate that the benefits of human-derived visual foresight depend on task characteristics, even when contact and action supervision remain robot-specific.

\subsection{Qualitative Analysis and Real-World Deployment}
\label{sec:qualitative_realworld}

\begin{figure}[t]
    \centering
    \includegraphics[width=\columnwidth]
    {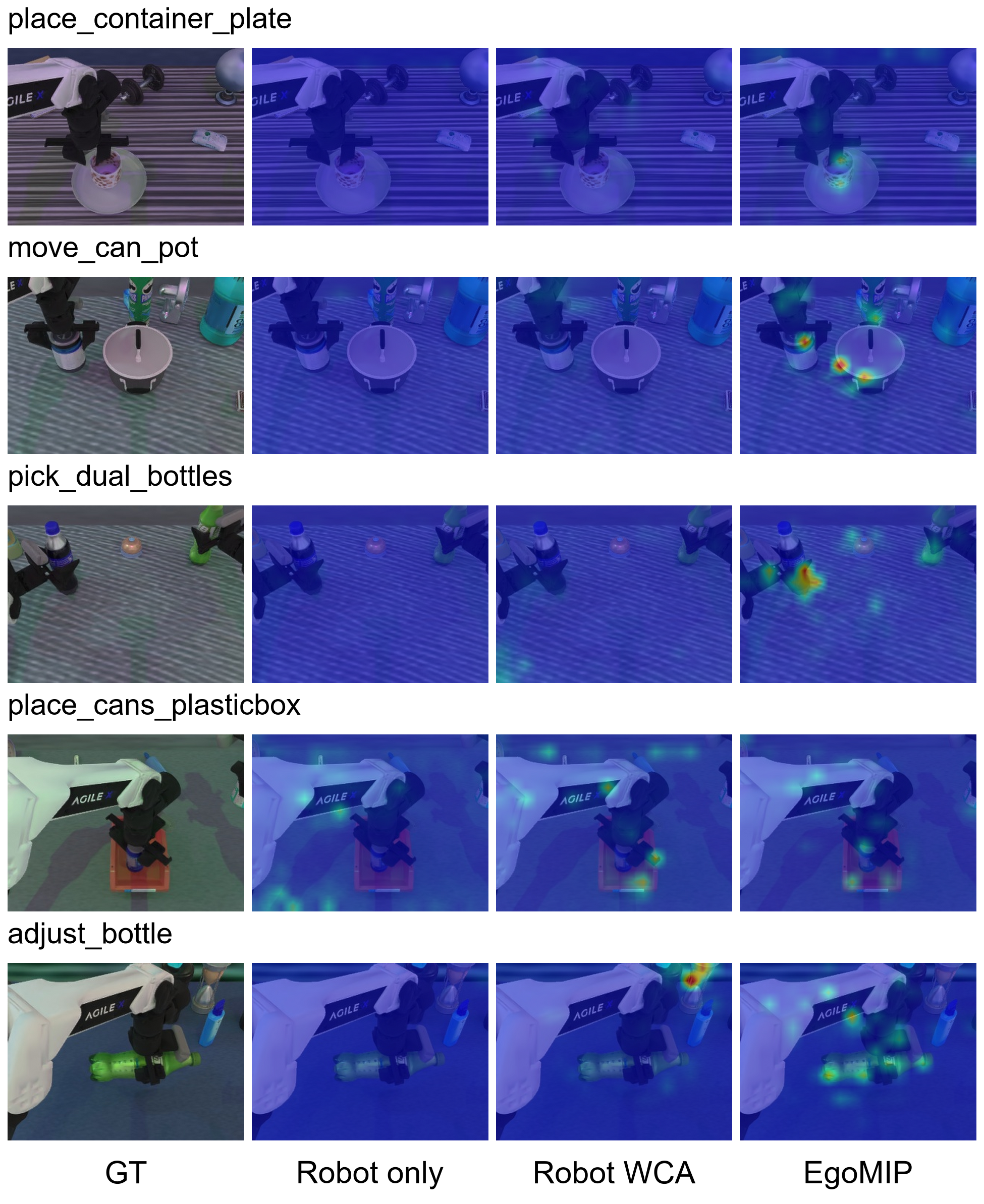}
    \vspace{-2mm}
    \caption{
        Attention visualization across representative
        RoboTwin tasks under different supervision configurations.
    }
    \label{fig:attention_visualization}
    \vspace{-2mm}
\end{figure}

\begin{figure}[t]
    \centering
    \includegraphics[width=\columnwidth]
    {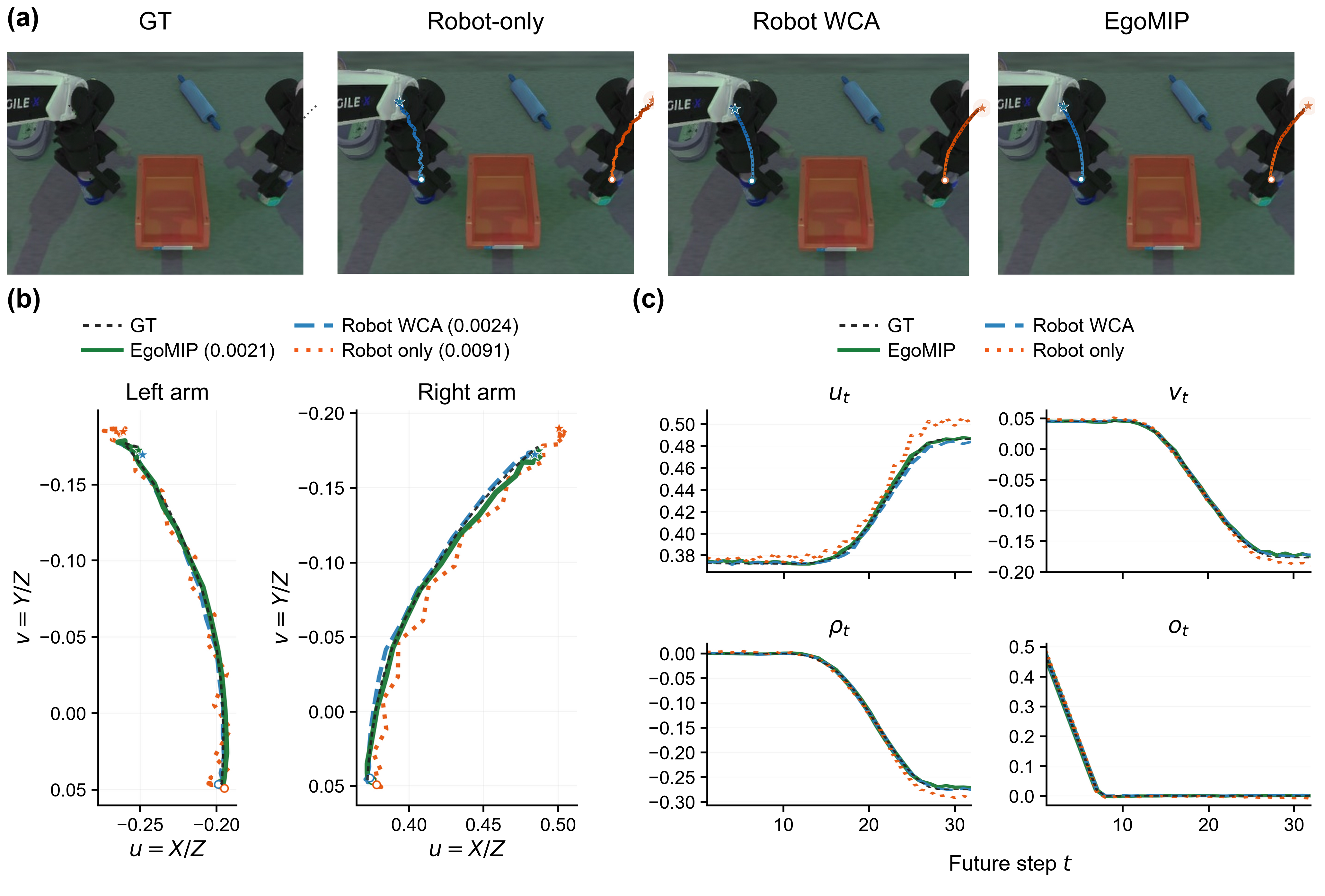}
    \vspace{-2mm}
    \caption{
        Contact trajectory analysis.
        (a) Scene-level contact visualization.
        (b) Projected TCP trajectories and mean $u$-$v$ errors.
        (c) Temporal evolution of 2.5D contact states.
    }
    \label{fig:contact_trajectory_analysis}
    \vspace{-2mm}
\end{figure}

\begin{figure}[t]
    \centering
    \includegraphics[width=\columnwidth]
    {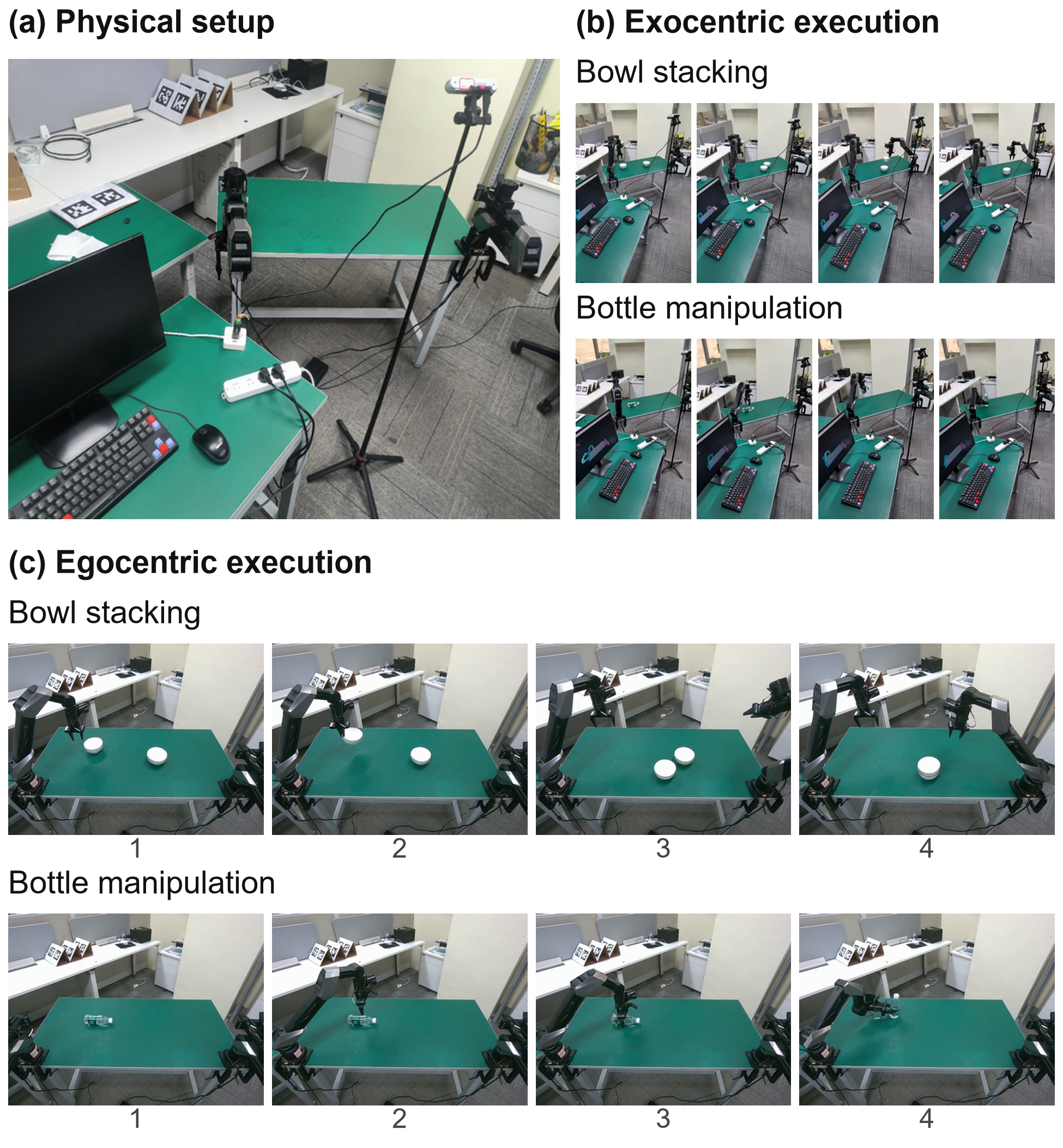}
    \vspace{-2mm}
    \caption{
        Real-world deployment of EgoMIP.
        (a) Physical dual-arm robot platform.
        (b) Exocentric execution of bowl stacking and bottle manipulation.
        (c) Egocentric execution of the corresponding tasks.
    }
    \label{fig:real_world_deployment}
    \vspace{-2mm}
\end{figure}
\noindent\textbf{Qualitative Analysis.}
To further examine the learned visual and interaction representations, we visualize the attention patterns of different supervision configurations across representative RoboTwin tasks. As shown in Fig.~\ref{fig:attention_visualization}, EgoMIP exhibits task-relevant attention responses around manipulated objects and their associated interaction regions. These visualizations provide qualitative evidence of how different supervision strategies shape the visual representations learned by the policy.

We further visualize the predicted 2.5D contact trajectories in Fig.~\ref{fig:contact_trajectory_analysis}. The results include scene-level bimanual contact paths, projected TCP trajectories, and the temporal evolution of contact states over 32 future steps. Compared with robot-only supervision, EgoMIP produces contact trajectories that more closely follow the ground-truth references in the illustrated examples, providing additional evidence for the role of structured contact modeling in connecting visual foresight with action generation. Table~\ref{tab:contact_trajectory_error} shows that EgoMIP achieves the lowest mean trajectory error ($2.258\times10^{-3}$), reducing it by 12.2\% over Robot WCA and 68.5\% over Robot-only. Additional contact trajectory analysis across ten representative tasks is provided in Appendix~\ref{sec:appendix_extended}.

\begin{table}[t]
    \centering
    \caption{
        Quantitative comparison of predicted 2.5D contact
        trajectories on two representative randomized
        RoboTwin 2.0 episodes. Lower values indicate
        better agreement with ground-truth TCP trajectories.
        Errors are averaged over both arms and reported
        in units of $10^{-3}$.
    }
    \vspace{-2mm}
    \label{tab:contact_trajectory_error}
    \small
    \begin{tabular*}{\columnwidth}{
        @{\extracolsep{\fill}}lccc@{}
    }
        \toprule
        Task
        & Robot-only
        & Robot WCA
        & EgoMIP \\
        \midrule
        Place Cans
        & 9.107
        & 2.394
        & \textbf{2.122} \\

        Stack Blocks Three
        & 5.214
        & 2.748
        & \textbf{2.394} \\

        \midrule
        Overall Mean
        & 7.161
        & 2.571
        & \textbf{2.258} \\
        \bottomrule
    \end{tabular*}
    \vspace{-6mm}
\end{table}

\noindent\textbf{Real-World Deployment.}
To examine physical feasibility, we deploy the
simulation-trained WCA policy on a physical dual-arm robot. A small set of teleoperated
demonstrations is used for task-specific post-training,
with 2.5D contact targets reconstructed from calibrated
camera extrinsics and forward kinematics. Fig.~\ref{fig:real_world_deployment} shows representative executions of bottle position adjustment
and Bowl Stacking. These results provide preliminary
evidence that the proposed policy can be adapted to
real-world dual-arm manipulation. Additional deployment
details and further quantitative evaluation are provided
in Appendix~\ref{sec:appendix_realworld}.

\section{Conclusion}

In this work, we presented EgoMIP, a unified World--Contact--Action framework for transferring interaction knowledge from human egocentric videos to robotic manipulation. By combining shared visual foresight with robot-specific 2.5D contact modeling and action learning, EgoMIP exploits complementary supervision across embodiments without requiring direct human-to-robot motion correspondence. Experiments on RoboTwin 2.0 demonstrate strong multi-task manipulation performance and improved robustness under scene randomization, while further analyses reveal that the benefits of human supervision depend on task characteristics and supervision allocation. Qualitative trajectory analysis and real-world dual-arm deployment further illustrate the practical potential of the proposed framework. These findings highlight the importance of selectively transferring visual knowledge while grounding physical interaction and control in robot demonstrations. Future work will explore longer-horizon task reasoning and additional robot-specific supervision to better handle complex interaction constraints and multi-stage manipulation.

\section*{LLM Usage Statement}

Large language models (LLMs), including Anthropic Claude and OpenAI
ChatGPT, were used exclusively to assist with preparing and refining
author-produced materials. Their use was limited to:
(1) improving the grammar, clarity, and organization of manuscript text; (2) editing author-drafted related work and standardizing citation formatting; (3) assisting with the refactoring, debugging, and documentation of
author-written code; (4) refining plotting and layout code, typography, and presentation of author-designed figures; and (5) providing reviewer-style feedback on manuscript clarity and presentation. No LLMs or generative AI tools were used to originate research ideas,
formulate hypotheses, establish scientific claims or conclusions,
design or conduct experiments, analyze experimental results, or generate
synthetic data, datasets, or annotations. All scientific contributions, experimental findings, and interpretations
are the authors' own. The authors reviewed and verified all AI-assisted
materials, take full responsibility for the submission, and can explain,
reproduce, and defend the work.
{
    \small
    \bibliographystyle{ieeenat_fullname}
    \bibliography{main}
}
\clearpage

\setcounter{section}{0}
\setcounter{subsection}{0}
\setcounter{figure}{0}
\setcounter{table}{0}
\setcounter{equation}{0}

\renewcommand{\thesection}{S\arabic{section}}
\renewcommand{\thesubsection}{\thesection.\arabic{subsection}}

\renewcommand{\thefigure}{S\arabic{figure}}
\renewcommand{\thetable}{S\arabic{table}}
\renewcommand{\theequation}{S\arabic{equation}}

\makeatletter
\@ifundefined{theHsection}{}{%
  \renewcommand*{\theHsection}{supp.\arabic{section}}}
\@ifundefined{theHsubsection}{}{%
  \renewcommand*{\theHsubsection}{supp.\arabic{section}.\arabic{subsection}}}
\@ifundefined{theHfigure}{}{%
  \renewcommand*{\theHfigure}{supp.\arabic{figure}}}
\@ifundefined{theHtable}{}{%
  \renewcommand*{\theHtable}{supp.\arabic{table}}}
\@ifundefined{theHequation}{}{%
  \renewcommand*{\theHequation}{supp.\arabic{equation}}}
\makeatother

\twocolumn[{
\begin{center}
    {\large\bfseries Appendix\par}
    \vspace{1em}
\end{center}
}]
\phantomsection
\addcontentsline{toc}{section}{Appendix}

\section{Real-World Robot Evaluation}
\label{sec:appendix_realworld}

\noindent\textbf{Setups.}
We evaluate EgoMIP on a Galaxea dual-arm robot
using two real-world tasks: Bowl Stacking and
Bottle Adjustment, as illustrated in
Fig.~\ref{fig:real_world_deployment} of the main paper.
The policy takes RGB observations and 14-D
joint proprioception as input and predicts
32-step action chunks.
Bowl Stacking requires bimanual coordination,
whereas Bottle Adjustment is performed
using the left arm.
All compared methods share identical visual
preprocessing, action interfaces, and control
configurations to ensure a fair comparison.

\noindent\textbf{Data Collection and Training.}
For each task, we collect 30 teleoperated
demonstrations for real-world adaptation.
We compare three variants: Robot-only W--A,
Robot WCA, and EgoMIP. All variants are
initialized from their corresponding
simulation-trained checkpoints and adapted
using identical real-world demonstrations
and post-training budgets. For contact-aware
variants, 2.5D contact targets are reconstructed
from calibrated camera extrinsics and robot
forward kinematics.

\noindent\textbf{Evaluation Results.}
We evaluate each policy over 30 autonomous trials
per task under matched initial configurations.
As shown in Table~\ref{tab:supp_real_sr},
EgoMIP achieves 83.3\% and 73.3\% success
on Bowl Stacking and Bottle Adjustment,
respectively, outperforming Robot WCA by
6.7 percentage points on both tasks.
The improvement from Robot-only W--A to
Robot WCA further supports the contribution
of explicit contact modeling.
These results provide preliminary evidence
of the physical applicability of EgoMIP,
although the limited trial count does not
establish statistically significant differences.

\begin{table}[t]
\centering
\small
\setlength{\tabcolsep}{4pt}
\begin{tabular}{lcc}
\hline
Method & Bowl Stacking & Bottle Adjustment \\
\hline
Robot-only W--A & 63.3 (19/30) & 56.7 (17/30) \\
Robot WCA       & 76.7 (23/30) & 66.7 (20/30) \\
\textbf{EgoMIP} & \textbf{83.3 (25/30)}
                & \textbf{73.3 (22/30)} \\
\hline
\end{tabular}
\caption{
Real-world manipulation success rates (\%).
Successful trials over total trials are
shown in parentheses. All variants use
identical real-world adaptation and
evaluation protocols.
}
\label{tab:supp_real_sr}
\end{table}

\section{Additional Ablation Studies}
\label{sec:appendix_ablations}

\noindent\textbf{Evaluation Protocol.}
Unless otherwise specified, all simulation
success rates reported in this section are
averaged over three independent training seeds
under the same RoboTwin 2.0 evaluation protocol.

\noindent\textbf{Contact Representation Ablation.}
We further investigate whether the improvements from
contact modeling arise from additional auxiliary
supervision or the geometric structure of the contact
representation. As shown in
Table~\ref{tab:supp_contact_ablation}, introducing a
non-geometric auxiliary prediction branch improves
performance over the W--A baseline, suggesting that
additional predictive supervision itself benefits
action learning. Explicit geometric modeling provides
further gains, with the 3D TCP representation achieving
92.41\% and 91.28\% success in clean and randomized
environments, respectively.

The proposed 2.5D representation further improves
performance to 93.06\% and 92.04\%, outperforming
3D TCP by 0.65 and 0.76 percentage points.
Unlike Cartesian 3D coordinates, the 2.5D formulation
combines normalized image-plane coordinates with
logarithmic depth and gripper openness. This
parameterization may facilitate the alignment of
contact trajectories with image-based visual
representations while providing a compact description
of interaction geometry. These results suggest that
both explicit geometric supervision and its
representation contribute to contact-aware action
learning, although the observed differences alone
do not establish the underlying mechanism.

\begin{table}[t]
\centering
\small
\setlength{\tabcolsep}{4pt}
\begin{tabular}{lcc}
\hline
Contact Representation & Clean & Rand. \\
\hline
None (W--A)              & 90.28 & 89.16 \\
Non-geometric Auxiliary  & 91.14 & 90.32 \\
3D TCP + Gripper         & 92.41 & 91.28 \\
2.5D TCP + Gripper (Ours)
                         & \textbf{93.06}
                         & \textbf{92.04} \\
\hline
\end{tabular}
\caption{
Contact representation ablation on 50 RoboTwin 2.0
tasks. All values denote success rates (\%).
The proposed 2.5D representation achieves the
highest performance under both clean and
randomized settings.
}
\label{tab:supp_contact_ablation}
\end{table}

\noindent\textbf{Adaptive World Supervision.}
We investigate the contribution of adaptive
action-effect feature weighting (AAEF) by
comparing it with uniform human World supervision.
Both configurations retain identical camera
alignment and agent masking, differing only in
the weighting of valid visual tokens.
As shown in Table~\ref{tab:supp_aaef},
AAEF improves success rates by 0.25 and 0.82
percentage points in clean and randomized
environments, respectively.

While uniform weighting treats all valid
visual regions equally, AAEF emphasizes
regions exhibiting larger future feature
changes. This encourages the World expert
to focus more strongly on potentially
interaction-relevant scene dynamics.
The larger improvement under randomization
is consistent with the intended role of
adaptive visual supervision, although
feature-change magnitude remains a heuristic
rather than a direct estimate of physical
action effects.

\begin{table}[t]
\centering
\small
\setlength{\tabcolsep}{6pt}
\begin{tabular}{lcc}
\hline
Human World Supervision & Clean & Rand. \\
\hline
Uniform Weighting & 93.11 & 92.52 \\
AAEF (Ours) & \textbf{93.36} & \textbf{93.34} \\
\hline
\end{tabular}
\caption{
Effect of adaptive action-effect feature
weighting on 50 RoboTwin 2.0 tasks.
Success rates (\%) are averaged over three
independent training seeds.
Both configurations use identical data,
agent masking, camera alignment, and
optimization settings, differing only in
human World loss weighting.
}
\label{tab:supp_aaef}
\end{table}


\section{Extended Experimental Results}
\label{sec:appendix_extended}


\noindent\textbf{Extended Contact Trajectory Analysis.}
To complement the representative examples in
Table~\ref{tab:contact_trajectory_error} of the main paper, we further evaluate
contact trajectory prediction on ten representative
contact-centric RoboTwin 2.0 tasks under randomized
environments. We compare Robot-only, Robot WCA,
and EgoMIP using the same trajectory error metric.

\begin{figure}[t]
    \centering
    \includegraphics[width=0.95\columnwidth]
    {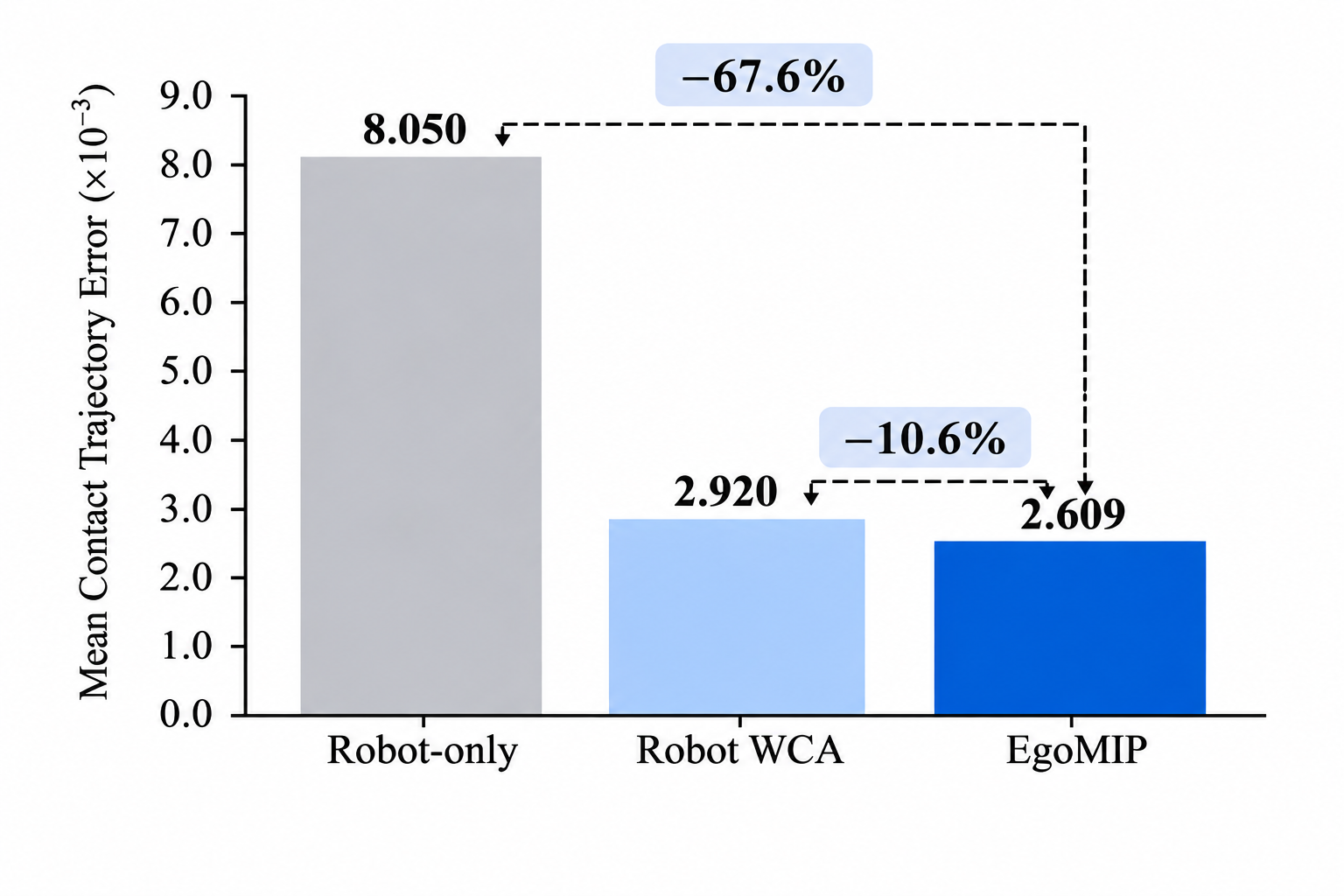}
    \caption{
    Mean contact trajectory errors over ten
    representative contact-centric RoboTwin 2.0
    tasks under randomized environments.
    Lower is better.
    }
    \label{fig:supp_contact_mean}
\end{figure}

As shown in Fig.~\ref{fig:supp_contact_mean},
the three methods achieve mean trajectory errors
of $8.050\times10^{-3}$, $2.920\times10^{-3}$,
and $2.609\times10^{-3}$, respectively.
EgoMIP reduces the average trajectory error
by approximately 10.6\% compared with Robot WCA
and 67.6\% compared with Robot-only.

These results further support the role of
structured contact modeling in improving
interaction prediction. The additional reduction
achieved by EgoMIP also suggests that human-derived
visual foresight provides complementary information
for robot-specific contact prediction.

\end{document}